\documentclass[10pt,twocolumn,letterpaper]{article}

\usepackage[pagenumbers]{wacv} %

\usepackage{graphicx}
\usepackage{amsmath}
\usepackage{amssymb}
\usepackage{booktabs}
\usepackage{multirow}
\usepackage[accsupp]{axessibility}

\usepackage[pagebackref,breaklinks,colorlinks]{hyperref}

\usepackage[capitalize]{cleveref}
\crefname{section}{Sec.}{Secs.}
\Crefname{section}{Section}{Sections}
\Crefname{table}{Table}{Tables}
\crefname{table}{Tab.}{Tabs.}

\def\wacvPaperID{56} %
\def\confName{WACV}
\def\confYear{2024}

\begin{document}
\newcommand{\rv}[1]{{\MakeUppercase{#1}}}  %
\newcommand{\set}[1]{\mathcal{{\MakeUppercase{#1}}}}  %
\newcommand{\vecspace}[1]{\mathbb{{\MakeUppercase{#1}}}} %

\newcommand{\argmax}{\operatornamewithlimits{argmax}} %
\newcommand{\argmin}{\operatornamewithlimits{argmin}} %

\newcommand{\dnet}{\ensuremath{D_{\theta}}}
\newcommand{\gnet}{\ensuremath{G_{\psi}}}
\newcommand{\pd}{\ensuremath{P_{\dnet}}}
\newcommand{\pg}{\ensuremath{P_{\gnet}}}
\newcommand{\pdata}{\ensuremath{P_{\text{data}}}}
\newcommand{\nsamples}{\ensuremath{S}}
\newcommand{\ndims}{\ensuremath{n}}
\newcommand{\zdims}{\ensuremath{d}}
\newcommand{\auxdims}{\ensuremath{m}}
\newcommand{\ld}{\ensuremath{\lambda_{\psi}}}

\newcommand{\FLOW}{flow}
\newcommand{\flow}{\FLOW}
\newcommand{\modelname}{our model}
\newcommand{\mytitle}{Adversarial Likelihood Estimation With One-Way Flows}

\newcommand{\qheading}[1]{\noindent\textbf{#1}}
\newcommand{\Qheading}[1]{\qheading{#1}}

\newcommand{\eqationref}[1]{Eq.~{\ref{#1}}}  %
\newcommand{\figref}[1]{figure~{\ref{#1}}}  %
\newcommand{\tableref}[1]{table~{\ref{#1}}}  %
\newcommand{\secref}[1]{Section~{\ref{#1}}}  %
\newcommand{\chapref}[1]{chapter~{\ref{#1}}}  %

\newcommand{\todoinline}[2]{\textcolor{red}{TODO\{#1\}:} \textcolor{blue}{#2}}

\newcommand{\lossreal}{\mathcal{L}_{real}}
\newcommand{\lreal}{\lossreal}

\newcommand{\lossgen}{\mathcal{L}_{fake}}
\newcommand{\lgen}{\lossgen}

\newcommand{\nn}{nn}
\newcommand{\gnety}{gnety}

\newcommand{\real}[1]{\mathbb{\mathbb{R}}^{#1}}

\title{\mytitle{}}

\author{Omri Ben-Dov$^{1}$,\enspace Pravir Singh Gupta$^{2}$,\enspace Victoria Abrevaya$^{1}$,\enspace Michael J. Black$^{1}$,\enspace Partha Ghosh$^{1}$\\
$^{1}$Max Planck Institute for Intelligent Systems, Tübingen, Germany \quad $^{2}$Perceive Inc.\\
{\tt\small \{odov, vabrevaya, black, pghosh\}@tuebingen.mpg.de, pravir.singh.gupta@gmail.com}
}

\maketitle

\begin{abstract}

Generative Adversarial Networks (GANs) can produce high-quality samples, but do not provide an estimate of the probability density around the samples.
However, it has been noted that maximizing the log-likelihood within an energy-based setting can lead to an adversarial framework where the discriminator provides unnormalized density (often called energy).
We further develop this perspective, incorporate importance sampling, and show that 1) Wasserstein GAN performs a biased estimate of the partition function, and we propose instead to use an unbiased estimator; and 2) when optimizing for likelihood, one must maximize generator entropy.
This is hypothesized to provide a better mode coverage.
Different from previous works, we explicitly compute the density of the generated samples.
This is the key enabler to designing an unbiased estimator of the partition function and computation of the generator entropy term.
The generator density is obtained via a new type of flow network, called one-way flow network, that is less constrained in terms of architecture, as it does not require a tractable inverse function.
Our experimental results show that our method converges faster, produces comparable sample quality to GANs with similar architecture, successfully avoids over-fitting to commonly used datasets and produces smooth low-dimensional latent representations of the training data.
\end{abstract}

\section{Introduction}
\label{sec:intro}

The goal of a generative model is 
to extract some notion of the data distribution given a training set, either explicitly by computing the probability density function~\cite{van2016pixel}, indirectly through distilling a stochastic sampling mechanism~~\cite{Goodfellow2014_GAN_GAN}, or a combination of both~\cite{dinh2016density}. 
While indirect generative models can achieve state-of-the-art performance in sample quality~\cite{Goodfellow2014_GAN_GAN, karras2019style}, having explicit densities  %
has several advantages.
For example, an explicit density function can be used to quantitatively compare models, 
or to 
train models by maximum likelihood estimation (MLE), which has been proven to be statistically asymptotically efficient~\cite{Huber67}.

Autoregressive models~\cite{van2016conditional, van2016pixel} and normalizing flows~\cite{dinh2014nice} are the most prominent examples of deep generative models that compute exact probability and directly maximize the log-likelihood of their training dataset. 
However, it is inefficient to sample from autoregressive models and they do not provide a low-dimensional latent representation of the data. 
Normalizing flows allow both efficient sampling and density estimation, but make restrictive assumptions on the architecture, requiring the latent space to be of the same dimensionality as that of the input, 
making it computationally expensive to use in a high-dimensional data regime.

Energy-based models (EBMs)~\cite{teh2003energy}, variational autoencoders (VAEs)~\cite{Kingma2014} and diffusion models~\cite{sohl2015deep, rombach2022high} %
are further examples of deep generative models trained with likelihood maximization. However, VAEs and diffusion models can only compute a lower bound of the likelihood. %
EBMs, on the other hand, represent an unnormalized density, allowing for greater flexibility in the choice of functional form, at the cost of inefficient sampling and approximate likelihood estimation.

Indirect models such as Generative Adversarial Networks (GANs)~\cite{Goodfellow2014_GAN_GAN, karras2019style} have achieved state-of-the-art performance in terms of the quality of the generated data, but do not provide any estimate of the probability density around a sample. However, a connection has been noted between the loss function of these networks, in particular the Wasserstein GAN (WGAN) loss~\cite{wasserstein_gan} and EBMs~\cite{xie16convnet,kim2016deep,dai2017calibrating}, in which the discriminator can be regarded as an energy function. With the goal of introducing density estimation within an adversarial training framework, we follow here a similar path, but develop further these observations to arrive at an \emph{unbiased} estimator of the partition function through the explicit computation of the generator density. 

Specifically, we begin by exploring the connection between EBMs and GANs which leads to  a training objective that closely resembles the WGAN loss, with minor but key differences. 
We notice that maximizing the log-likelihood of an EBM arrives at the WGAN loss if we take a biased estimation of the normalization constant of the energy function; or alternatively, WGANs perform a one-sample approximation of the partition function. Based on this observation, and in departure from previous work, we propose to use an unbiased estimator by explicitly computing the generator density $\pg$. 

To calculate $\pg$, we propose a new type of normalizing flow network that bypasses several architectural constraints found in standard flow models. In particular, we construct %
a flow that can perform upsampling and downsampling operations, starting from a lower-dimensional latent variable, at the cost of approximate probability computation. This is possible and sufficient since we only need to compute $\pg$ for generated samples, while density estimation of real, non-generated points is relegated to the discriminator.

Our experimental results show that our model is able to capture more modes, trains faster on images, produces comparable sample quality to GANs with similar architecture,  and can be used to compute the partition function
with a practical number of samples.

In summary, we propose a framework for adversarial generative modeling that simultaneously computes an estimate of the density, with the following key contributions: i) by developing the connection between EBMs and GANs, we show that the WGAN discriminator objective is a biased estimator of the partition function; ii) we propose an unbiased estimate of the partition function of an EBM by explicitly computing the density of the generator; iii) we propose a new flow-based network for the computation of the generator density that enables a more flexible architecture, in contrast to traditional flow models.

\section{Related work}
\label{sec:related_work}

Two main categories of generative models are \emph{prescribed} and \emph{implicit} models~\cite{Diggle84}. Prescribed models recover an explicit parametric specification of the density function and are trained and evaluated by MLE; our work belongs to this family. 
Implicit models, on the other hand, represent the data distribution indirectly through a stochastic mechanism that generates random samples. In general, this offers more flexibility in terms of learning objective and model architecture, which is hypothesized to be responsible for the high visual quality of the generated samples. 

Normalizing flows~\cite{kobyzev2020normalizing, dinh2016density, Kingma2018_glow} and autoregressive models~\cite{theis2015generative, van2016pixel, salimans2017pixelcnn++} are examples of deep \emph{prescribed} generative models. 
Since these compute the density function explicitly, they can be optimized and evaluated using the train- and test-set log-likelihood. Although autoregressive models can efficiently work with high-dimensional data during training, due to ancestral sampling they are extremely slow at generating new samples. Normalizing flows require an invertible architecture to compute the likelihood, and consequently can only  support latent spaces of the same dimensionality as the input data. In addition, they tend to produce large and memory-hungry models, and are therefore not so suitable for high-dimensional data. 
In this work, we relax the invertibility constraint by computing the flow in only one direction, enabling the use of lower-dimensional latent vectors, and more resource-efficient architectures. 

An intermediate category of generative models considers only an \emph{approximation} to the density function.
Examples include a lower bound on the likelihood for VAEs and diffusion models~\cite{ho2020denoising}, or the unnormalized density in the case of EBMs~\cite{song2021train}. VAEs are known to suffer from low generation quality, \ie they tend to produce blurry samples. Diffusion models can generate images of very high sample-quality~\cite{dhariwal2021diffusion, rombach2022high}; however, the latent representation needs to be of the same dimension as the input data. EBMs~\cite{teh2003energy} deploy several techniques to obtain the derivative of the normalizing factor with respect to the model parameters. We maximize the same cost function as EBMs (see \cref{eq:disc_objective}), but explicitly model the normalization constant $\zeta$. 

GANs~\cite{Goodfellow2014_GAN_GAN} are the most prominent example of \emph{implicit} models, and produce state-of-the-art generated sample quality~\cite{karras2019analyzing}. However, it has been observed that GANs may trade diversity for precision~\cite{Che16, Salimans16, srivastava2017veegan}. This results in generators that produce samples from only a few modes of the data distribution, a phenomenon known as ``mode collapse''.
GANs are also well known for having unstable training dynamics~\cite{wasserstein_gan, mescheder2018training, heusel2017gans}. 

The connection between GANs and EBMs on which we base our analysis has been previously observed in \cite{xie16convnet, kim2016deep, dai2017calibrating, che2020your}.
A common assumption by these works is that the generator density is inaccessible, which forces them to work with a biased partition function.
Furthermore, while designing the objective for the generator network, it is observed (as in this work) that the entropy of the generator distribution needs to be maximized.
However, entropy estimation is closely related to density estimation, and therefore as hard as the original problem. 
To address this, \cite{kim2016deep} assumes that the batch normalization layer maps every intermediate activation to approximately normal distributions, and that the sum of the analytical entropy of these distributions approximate the true generator entropy. 
In \cite{dai2017calibrating} two different approaches are proposed for the generator distribution: 1) assume that the  distribution is a mixture of isotropic Gaussians centered around the generator response $\gnet(z)$, and compute the gradient of such a mixture; and 2) compute its variational lower bound, which requires training yet another network that outputs a parametric form of the approximate posterior and an MCMC integration over the noise variable.
A similar approach using variational lower bound has also been explored in \cite{abbasnejad2019generative}.
Contrary to these, we explicitly compute the generator density and with it the entropy term, resulting in an unbiased estimate of the partition function.

\section{Method}
\label{sec:method}

\subsection{Density estimation by MLE}
\label{ssec:model_def}

Given a dataset of independent and identically distributed samples $\set{x} {:=} \{x_i \in \vecspace{R}^n\}_{i=1}^{m}$, drawn from an unknown probability distribution $\pdata$, our goal is to learn a parametric model $\pd$ that matches the distribution of $\pdata$.
Following EBMs~\cite{teh2003energy, song2021train}, we define $\pd$ as
\begin{equation}
\label{eq:pd}
    \pd \left(x\right)=\frac{e^{\dnet\left(x\right)}}{\zeta}.
\end{equation}
Here, $\dnet$ is a neural network with parameters $\theta$. The exponentiation ensures a non-negative probability, and $\zeta{=}\int_{\mathbb{R}^{\ndims}}e^{\dnet\left(x\right)}\text{d}x$ is a normalizing factor such that $\pd$ integrates to unity. Note that traditionally the energy of an EBM is represented as $e^{-D_\theta(x)}$, but here we consume the negative sign inside the $D_\theta(x)$ function for notational simplicity.

Since the partition function $\zeta$ is generally intractable and hard to compute, %
we approximate this integral with importance sampling~\cite{kloek1978bayesian}, and rewrite it as
\begin{equation}
\label{eq:integral_approx}
    \begin{split}\zeta & =\int_{\mathbb{R}^{\ndims}}\pg\left(x\right)\frac{e^{\dnet\left(x\right)}}{\pg\left(x\right)}\text{d}x=\mathbb{E}_{x\sim\pg}\left[\frac{e^{\dnet\left(x\right)}}{\pg\left(x\right)}\right]\\
 & \approx\frac{1}{\nsamples}\sum_{x\sim\pg}\frac{e^{\dnet\left(x\right)}}{\pg\left(x\right)},
\end{split}
\end{equation}
with $\nsamples$ representing the number of samples used in the summation, 
and where $\pg$ is an arbitrary distribution that is non-zero in the integration domain, often called the biased density. 
Here, we choose $\pg$ to be the push-forward density through a neural network $\gnet:\mathbb{\mathbb{R}}^{d} \rightarrow\mathbb{\mathbb{R}}^{n}$ with parameters $\psi$, such that $y = \gnet\left(z\right)$ is a sample from the biased density $\pg$, and $z \in \mathbb{R}^{\zdims}$, $z \sim P(Z)$ is the latent random variable, with $d \leq n$.  
We further choose $P(Z)$ to be the standard normal density $\mathcal{N}(0, I)$.
We will elaborate more on $\pg$ in~\cref{ssec:gen_prob}, and on $\gnet$ in~\cref{ssec:arch}.

We train $\dnet$ by maximizing the log-likelihood $\log P_{D}\left(x\right)$ of the dataset $\set{x}$: 

\begin{equation}
\label{eq:disc_objective}
    \begin{split}\theta^{*} & =\arg\max_{\theta}\left\{ \sum_{x\in\mathcal{X}}\log \pd \left(x\right)\right\} \\
 & =\arg\max_{\theta}\left\{ \sum_{x\in\mathcal{X}}\left(D_{\theta}\left(x\right)-\log\zeta\right)\right\} \\
 & \approx \arg\max_{\theta}\left\{ \sum_{x\in\mathcal{X}}\left[D_{\theta}\left(x\right)-\log\sum_{y\sim\pg}\frac{e^{\dnet\left(y\right)}}{\pg\left(y\right)}
 \right]\right\}.
 \end{split}
\end{equation}

The summation over $y\sim\pg$ is the $\zeta$ integral approximation from \cref{eq:integral_approx}, summed over $S$ samples. A full derivation can be found in \cref{ssec:disc_derivation}.
Interestingly, if we take a one-sample approximation %
we get the objective
\begin{equation}
\label{eq:disc_objective_wgan}
    \begin{split}\theta^{*} & =\arg\max_{\theta}\left\{ \sum_{x\in\mathbb{\mathcal{X}}}\left[\dnet\left(x\right)-\log e^{\dnet\left(y\right)}+\log\pg\left(y\right)\right]\right\} \\
 & =\arg\max_{\theta}\left\{ \sum_{x\in\mathbb{\mathcal{X}}}\left[\dnet\left(x\right)-\dnet\left(\gnet\left(z\right)\right)\right]\right\}.
\end{split}
\end{equation}
Here the term $\log\pg\left(y\right)$ can be discarded because it does not depend on $\theta$. \cref{eq:disc_objective_wgan} is exactly the objective for the WGAN discriminator~\cite{wasserstein_gan}. 
Hence, we note a connection between the unnormalized log-density estimator $\dnet$ and the WGAN objective function, 
which can be re-interpreted as performing a one-sample approximation of $\zeta$ within an energy-based framework.

There is an alternative view to \cref{eq:disc_objective_wgan}. %
If we simply drop the importance sampling scheme and embrace a biased estimate of the partition function, we again recover the WGAN objective as was shown in \cite{dai2017calibrating}.
The introduction of the importance weights as in \cref{eq:disc_objective} is known to produce an unbiased estimator \cite{rubinstein2016simulation} of the normalizing constant $\zeta$, although it does so at the cost of added variance. 
Since it is intractable to theoretically compute the variance of this estimator, even when we have access to the variance of the importance weight, we will show the empirical relevance of the unbiased estimator in~\cref{LED:sec:expt}.

\subsection{Learning $\pg$ for importance sampling}
\label{ssec:gen_loss}

The construction of $\pg$ %
in~\cref{eq:integral_approx,eq:disc_objective} is important, since an appropriate choice can dramatically reduce the number of samples required to achieve an accurate approximation of $\zeta$.

To reduce the number of samples needed we minimize the variance of the approximation error, which is proportional to $\frac{\pd}{\pg}$~\cite{rubinstein2016simulation}.
This occurs when $\pg$ matches ${\pd}$ up to a multiplicative factor. 
Therefore, we train $\gnet$ by minimizing the KL-divergence \cite{kullback1951information} between the two distributions, leading to the objective function:
\begin{equation}
\label{eq:gen_objective}
    \psi^{*}=\arg\max_{\psi}\left\{ H\left(\gnet\left(Z\right)\right)+\frac{1}{m}\sum_{z\sim Z}\dnet\left(\gnet\left(z\right)\right)\right\} ,
\end{equation}
where $Z$  is a random variable that is used as input to $\gnet$ and $H\left(\gnet\left(Z\right)\right)$ is the entropy of the generator distribution. 
The full derivation is in \cref{ssec:gen_derivation}.
\footnote{In practice, the entropy $H\left(\gnet\left(Z\right)\right)$ is not the same order of magnitude as the discriminator response, hence we add a weight $w$ to the entropy term. We mathematically justify this and correct the objective and probability using this weight in \cref{sec:jacobian_weight}.}

Notably, we obtain in~\cref{eq:gen_objective} the WGAN \emph{generator} objective, with an additional entropy term $H\left(\gnet\left(Z\right)\right)$ that requires maximization. We hypothesize that this term is responsible for ensuring diversity in the generated samples, and that its introduction can reduce the well-known problem of mode collapse in GANs. This has also been observed in~\cite{kim2016deep, dai2017calibrating}, where the authors proposed ad-hoc solutions to the computation of $H\left(\gnet\left(Z\right)\right)$ since the distribution of $\pg$ was unknown. 

Note that any other choice of divergence is in principle valid as objective function. We take here the KL-divergence because it leads to an objective for $\gnet$ that is independent of the normalizing constant $\zeta$ of the distribution given by $\dnet$, and because it leads to a natural connection with the WGAN loss. We leave the exploration of other divergences for future work.

\subsection{Estimating probabilities for the generator}
\label{ssec:gen_prob}

We require a tractable $\pg (y)$ for the approximation of the integral in~\cref{eq:disc_objective}.
One design option that fits this requirement is a normalizing flow network \cite{dinh2016density}, where the density at a point $y$ sampled using the generator network $\gnet (z)$ is computed using the change of variables formula: 
\begin{equation}
\label{eq:change_of_var}
    P\left(\gnet\left(z\right)\right)=P_{Z}\left(z\right)\left|\det\left(\frac{\partial \gnet\left(z\right)}{\partial z^{\intercal}}\right)\right|^{-1},
\end{equation}
with $P_{Z}$ the latent density from which $z$ is sampled, and $\frac{\partial \gnet\left(z\right)}{\partial z^{\intercal}}$ the Jacobian of $\gnet\left(z\right)$. 

Normalizing flows require the mapping $\gnet$ to be bijective for the change of variable formula given by~\cref{eq:change_of_var} to hold.
Additionally, for any $x \in \mathcal{X}$, normalizing flows must find a $z$ such that $\gnet\left(z\right) = x$.
This requires $\gnet$ to be designed in such a way that it can be efficiently inverted, which greatly restricts the choice of architecture of $\gnet$, and prevents from adopting the recent progress made by empirical research on GAN architectures~\cite{karras2019analyzing, reed2016generative}. %

In our setting, however, we need to evaluate the generator density only at points \emph{sampled from the generator} $y = \gnet(z)$. Therefore, we do not need to compute the inverse function $\gnet^{-1}(x)$ explicitly, only the forward $\gnet(z)$ and its Jacobian determinant. This allows to use any architecture for $\gnet$ whose Jacobian determinant can be computed efficiently. In Section~\ref{ssec:arch} we show how to build such architecture, which we call \emph{one-way flow}. 

\subsection{One-way flow generator network}
\label{ssec:arch}
Motivated by the generation quality of GANs we design a generator that maps a latent space ($\mathbb{R}^d$) to the data space ($\mathbb{R}^n$) with $d \ll n$, gradually increasing dimensionality while retaining  computational efficiency in the estimation of the density.

First, we define a function $g_{u}:\mathbb{R}^{d}\rightarrow\mathbb{R}^{n}$
that increases dimensionality by concatenating a random vector
$r\in\mathbb{R}^{n-d}$ as $g_u\left(z\right)=\begin{pmatrix}z\\
r
\end{pmatrix}$.
Since $r$ and $z$ are independent by design and since $P\left(r\right)$ and $P\left(z\right)$ are known, the probability of the output is 
\begin{equation}
\label{eq:random_concat}
    P\left(g_{u}\left(z\right)\right)=P\left(z\right)P\left(r\right).
\end{equation}

We can compose any number of functions that are either bijective or concatenate random noise as in $g_u$, although in practice we did not find a need to use more than one such layer.
We encapsulate the subsequent layers into a function of the form $g_{n}:\real{n}\rightarrow\real{n}$.
For the encapsulated layers, we allow any architecture design, with the constraint that nowhere inside this part of the network the dimension of the activation be smaller than $n$.

Finally, we construct the generator as $\gnet\left(z\right){=}\left(g_{n}\circ g_{u}\right)\left(z\right)$.
The form $\gnet:\mathbb{R}^{d}{\rightarrow}\mathbb{R}^{n}$ makes it possible to compute the probability $\pg\left(\gnet\left(z\right)\right)$ of an $n$-dimensional sample $\gnet\left(z\right)$ on its corresponding $d$-dimensional point $z$ on the manifold.
It is unnecessary to compute $\pg\left(x\right)$ on any arbitrary $n$-dimensional point $x$, since our model requires only probabilities of generated samples.

However, computing the Jacobian of $g_n$ and its determinant in high dimensions is a computationally heavy task.
To efficiently approximate the determinant of the Jacobian we use the equality
\begin{equation}
\label{eq:det_approx}
    \left|J\right|^{-1}=\mathbb{E}_{v\sim S^{n-1}}\left[\left\Vert Jv\right\Vert ^{-n}\right]
\end{equation}
from \cite{sohl2020two}, where $J \in \real{n \times n}$ is a matrix, \ie the Jacobian, and $v$ is a random unit vector.
To further increase efficiency, we use a one-sample approximation, which allows us to rewrite it in log form as
\begin{equation}
\label{eq:jacobian_approx}
    \log\left|J\right|\approx n\log\left\Vert Jv\right\Vert .
\end{equation}
We show in \cref{sec:optim_jacobian_effect} that using this form is sufficient for our purposes.

For our definition of $\gnet$, with \cref{eq:change_of_var,eq:random_concat}, we get the computationally efficient form of the generator density evaluated at a generated point as follows:
\begin{equation}
    \pg\left(\gnet\left(z\right)\right) = P\left(z\right)P\left(r\right)\left|\det\left(\frac{\partial g_{n}\left(z\right)}{\partial z^{\intercal}}\right)\right|^{-1}.
\end{equation}
Using the approximation of \cref{eq:jacobian_approx} we obtain a computationally efficient unbiased estimator of the entropy using an $m$-sample empirical mean as
\begin{equation}
    H\left(\gnet\left(z\right)\right)\approx-\frac{1}{m}\sum_{z\sim P_{Z}}\left[\log\left(P\left(z\right)P\left(r\right)\right)-n\log\left\Vert Jv\right\Vert \right].
\end{equation}
This in turn lets us write the generator objective as
\begin{equation}
    \psi^{*}=\arg\max_{\psi}\left\{ \sum_{z\sim P_{Z}}\left(n\log\left\Vert Jv\right\Vert +\dnet\left(\gnet\left(z\right)\right)\right)\right\}.
\end{equation}

The objective for the generator includes a maximization of the $\log$ determinant of the generator Jacobian.
This ensures that the Jacobian stays full rank during training. Furthermore, due to optimization dynamics, if it so happens that the Jacobian ceases to be full rank or approaches singularity, the cost function approaches negative infinity making the training dynamics shift and ``focus'' on restoring Jacobian rank.
In practice, we do not see the Jacobian approach singularity, and it stays well-behaved.

\section{Experiments}
\label{LED:sec:expt}

\begin{figure*}[t]
\begin{center}
\centerline{
\includegraphics[width=\columnwidth]{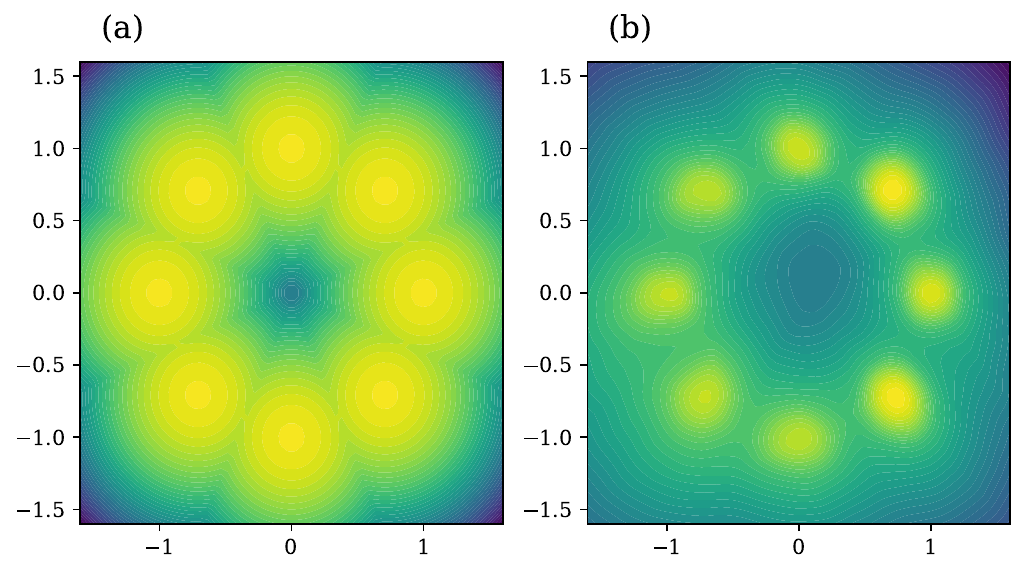}
\includegraphics[width=\columnwidth]{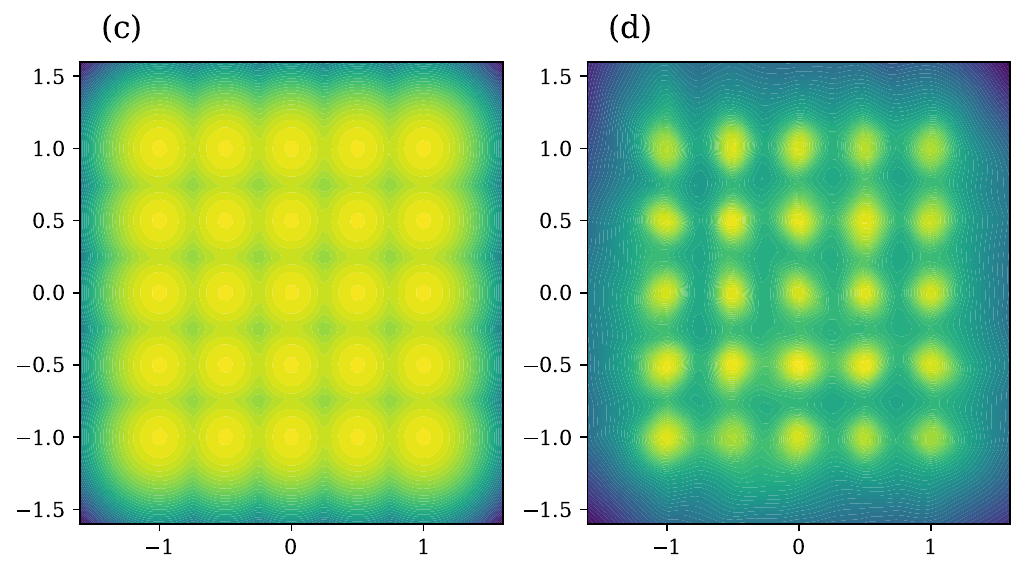}
}
\caption{True distribution v.s. discriminator distribution in log space. Brighter colors represent higher values. (a) Ground truth distribution of the 8-modes ring. (b) Discriminator density estimation of the 8-modes ring. (c) Ground truth distribution of the 25-modes grid. (d) Discriminator density estimation of the 25-modes grid.}
\label{fig:toy_dists}
\end{center}
\end{figure*}

\begin{figure*}
    \centering
    \includegraphics[height=0.5\columnwidth]{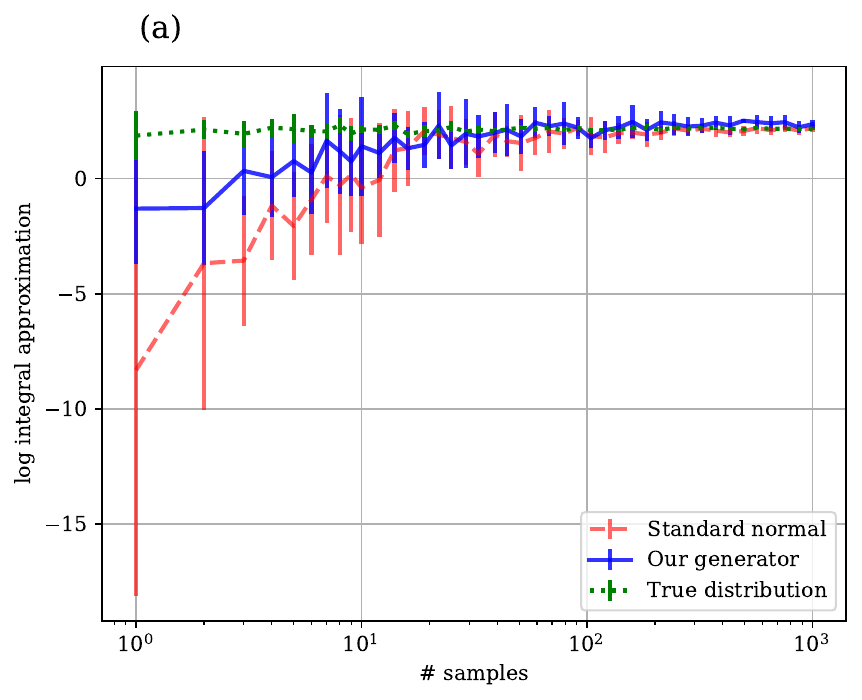}
    \includegraphics[height=0.5\columnwidth]{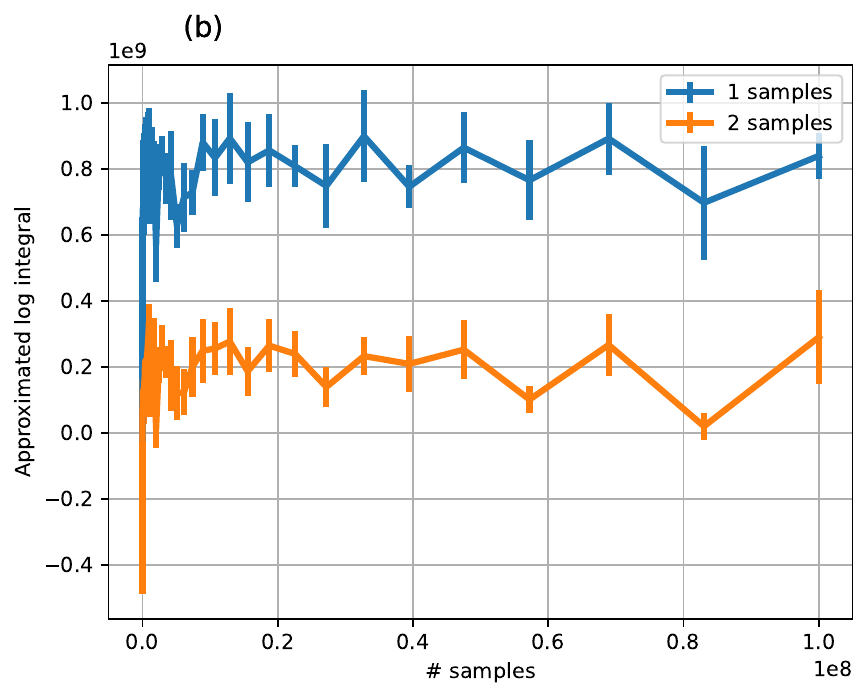}
    \includegraphics[height=0.5\columnwidth]{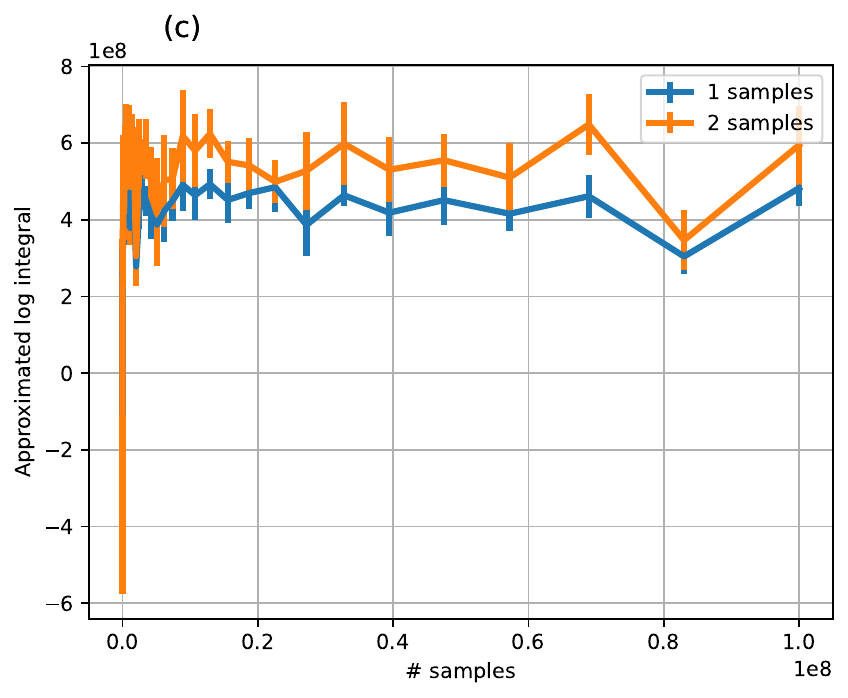}
    \caption{Approximated integral by number of samples used (as in \cref{eq:integral_approx}) for the grid distribution. The vertical error bars represent the standard deviation over 10 computations. (a) Synthetic 2D grid. (b) CelebA. (c) CIFAR-10.}
    \label{fig:toy_z_converge}
\end{figure*}

In this section we provide experimental results for the generated data and the density estimation using both synthetic (\cref{sec:results_synth_data}) and real (\cref{sec:results_real_data}) datasets.
We show qualitative examples in \cref{sec:results_qual}.
Implementation details and architectures can be found in \cref{sec:implemetation_dets} and \cref{sec:architecture}.

\subsection{Synthetic data} 
\label{sec:results_synth_data}

\begin{table}[!ht]
\caption{The high quality (HQ) samples percentages and modes of various generators for 2D GMMs. The numbers for all models, except our model, are taken from~\cite{srivastava2017veegan}. 
}
\label{table:2dgen_hq}
\vspace{0.1in}
\begin{center}
\begin{small}
\begin{sc}
\begin{tabular}{@{}lcccccc@{}}
\toprule
\multirow{2}{*}{}     & \multicolumn{2}{c}{\textbf{2D Ring}}   & \multicolumn{2}{c}{\textbf{2D Grid}}    \\  \cmidrule(r{1em}l{1em}){2-3} \cmidrule(r{1em}l{1em}){4-5}
                      & \textbf{\begin{tabular}[c]{@{}c@{}}Modes \\ 
                      (Max 8)\end{tabular}} & \textbf{\begin{tabular}[c]{@{}c@{}}\% HQ\\ 
                      \end{tabular}} & \textbf{\begin{tabular}[c]{@{}c@{}}Modes \\ 
                      (Max 25)\end{tabular}} & \textbf{\begin{tabular}[c]{@{}c@{}}\% HQ\\
                      \end{tabular}} \\ 
                      \midrule
\textbf{GAN~\cite{Goodfellow2014_GAN_GAN}}          & 1        & \textbf{99.3}            &3.3               &0.5           \\ 
\textbf{ALI~\cite{DumoulinBPLAMC17ALI}}          & 2.8      & 0.13        & 15.8         & 1.6          \\ 
\textbf{UGAN~\cite{MetzPPS17unrolled}} & 7.6      & 35.6    & 23.6        & 16            \\ 
\textbf{VEEGAN~\cite{srivastava2017veegan}}     & \textbf{8}       & 52.9         & 24.6         & 40               \\ 
\textbf{Ours}     &  \textbf{8}       & 71.2        & \textbf{25}         & \textbf{52.5}               \\ 
\bottomrule
\end{tabular}
\end{sc}
\end{small}
\end{center}
\vspace{0.1in}
\end{table}

We begin by comparing the density estimation and sampling capabilities of our model.
To this end, we perform an experiment on a synthetic 2D dataset, the same as the one presented in VEEGAN~\cite{srivastava2017veegan}.
We train our model on two sets of Gaussian Mixture Models (GMM), with one set comprising $8$ modes forming a ring (\cref{fig:toy_dists}a) and another set comprising $25$ modes in a grid (\cref{fig:toy_dists}c).
In order to test the theoretical analysis (\cref{sec:method}) without the consequences of approximating the determinant of the Jacobian (\cref{eq:jacobian_approx}), we performed the 2D experiments while computing the exact determinant of the $2 \times 2$ Jacobians.

To quantify the quality of the density captured by the \textbf{generator} we use the ``high-quality samples and modes'' metric from~\cite{srivastava2017veegan}, where a generated point is considered \emph{high quality} if it is within a $3\sigma$ distance from the nearest mode, and a mode is counted if it is the nearest mode to at least one high-quality sample.
We generate 2,500 points and report the percentage of points that are high quality and the number of modes over five runs.
We can see in \cref{table:2dgen_hq} that our generator is able to capture all the modes, while also producing higher quality samples than other models.

Since GAN models do not return a direct estimate of the probability of the data we cannot compare density estimation.
Therefore, instead of a quantitative comparison, we qualitatively evaluate the density estimation of our discriminator by plotting in \cref{fig:toy_dists} its density map next to the ground truth density.
\cref{fig:toy_dists} shows that our discriminator captures all the modes by giving them high values.

To show the effectiveness of the samples created by the generator, in \cref{fig:toy_z_converge}(a) we show the $\log \zeta$ approximated by a different number of samples using three different bias distributions: 1) a standard normal distribution, 2) our generator distribution and 3) the ground-truth distribution.
For each distribution and each number of samples, we run the computation 10 times, and use an error bar to represent the standard deviation of the results.
\cref{fig:toy_z_converge}(a) shows that using our generator is more accurate than  using a normal distribution, and requires fewer samples to converge.

\begin{figure}
    \centering
    \includegraphics[width=\columnwidth]{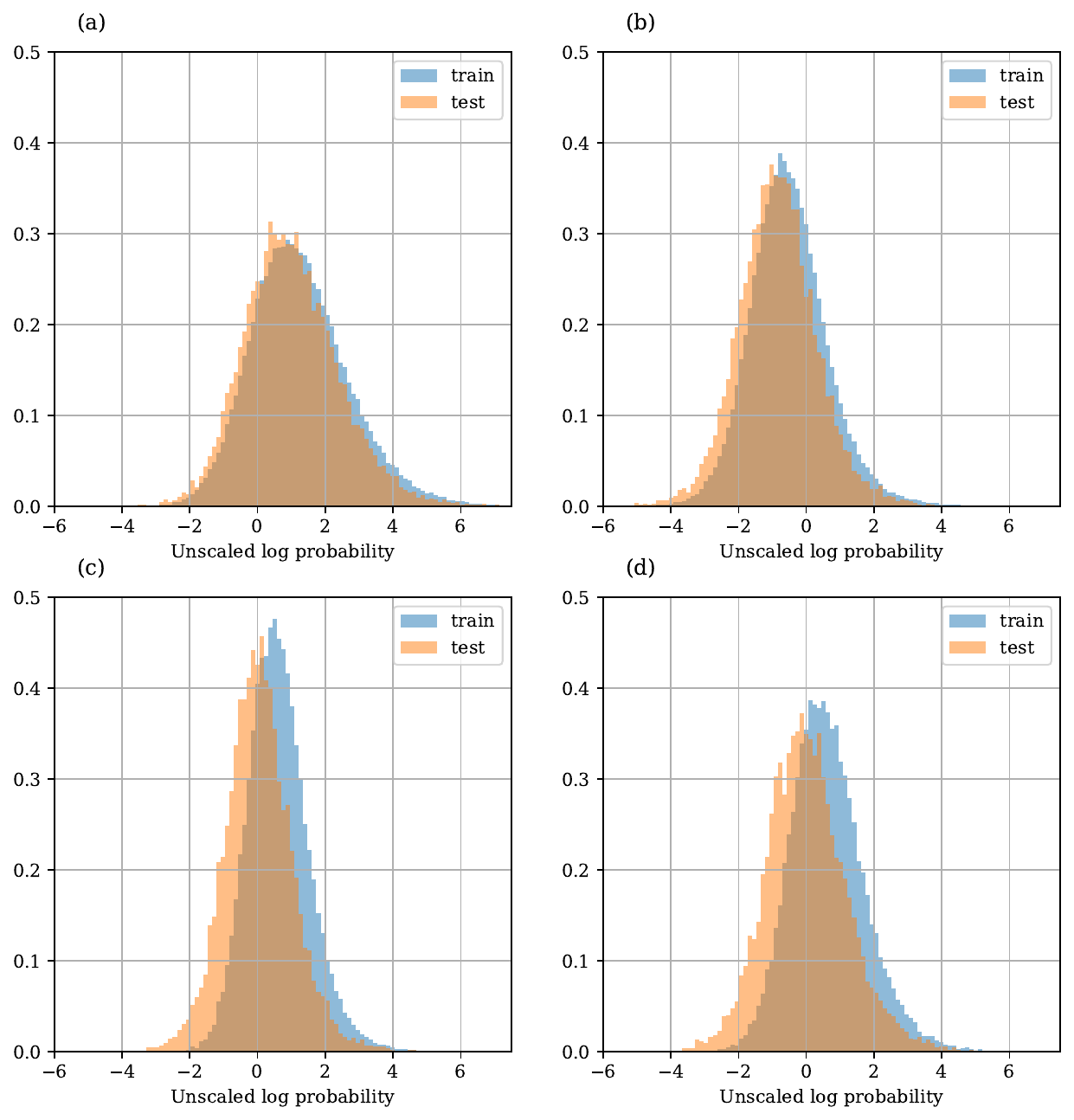}
    \caption{Overfit test. Histograms of the values returned by the discriminator for the train and test sets. Top row (a-b) for CelebA and bottom row (c-d) for CIFAR-10. The left column (a,c) uses a 1-sample approximation and $\zeta$ and the right column uses a 2-samples approximations.}
    \label{fig:overfit}
    \vspace{0.2in}
\end{figure}

\begin{figure}[t]
    \centering
    \includegraphics[width=\columnwidth]{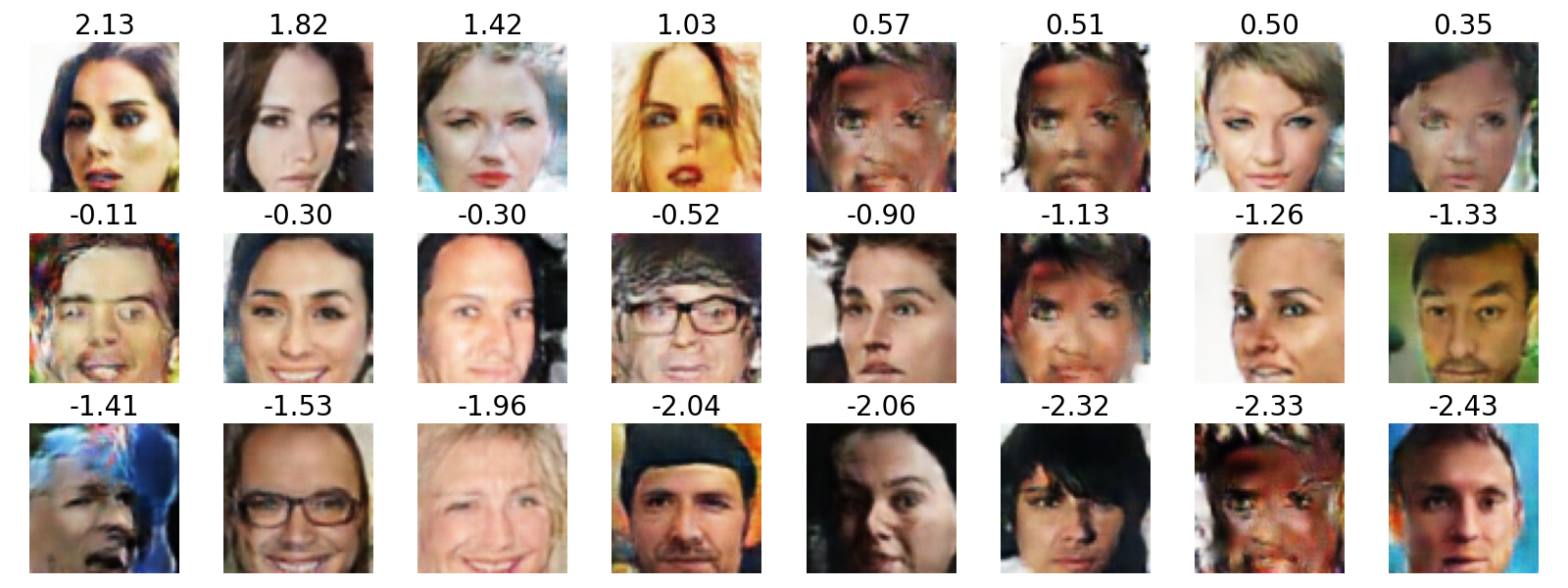}
    \caption{Random samples of generated CelebA images sorted by  their discriminator-assigned unnormalized log probability. The value above each image is the discriminator score.}%
    \label{fig:celeba_sort}
    \vspace{0.2in}
\end{figure}

\begin{figure}[t]
    \centering
    \includegraphics[width=\columnwidth]{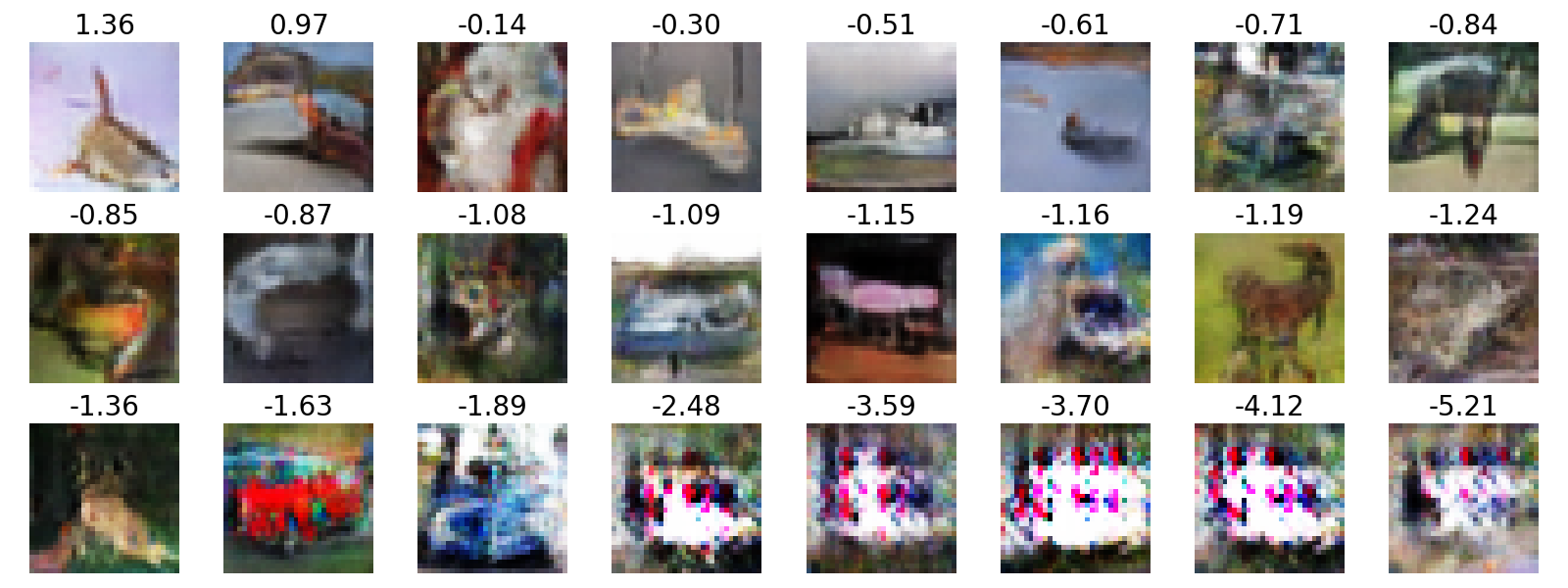}
    \caption{Random samples of generated CIFAR-10 images sorted by their discriminator-assigned unnormalized log-probability. The value above each image is the discriminator score.}%
    \label{fig:cifar_sort}
\end{figure}

\subsection{Real data}
\label{sec:results_real_data}

To evaluate our loss objectives (\cref{eq:disc_objective,eq:gen_objective}) on real datasets, we use the DCGAN~\cite{radford2015dcgan} architecture and train the model with various numbers of samples to approximate the integral (\cref{eq:integral_approx}).
\cref{sec:implemetation_dets} details our training parameters, \cref{sec:dcgan_arch} explains our DCGAN-based architecture and \cref{ssec:runtime} compares the runtime difference between WGAN and our method.
As seen in \cref{table:FID_score_b_ps}, using our formulation we achieve better Fréchet Inception Distance (FID) values.
The table also shows results from the traditional normalizing flow-based GLOW~\cite{Kingma2018_glow}.

Following \cite{du2019implicit}, we provide histograms of unnormalized log-likelihoods for train and test data in \cref{fig:overfit}.
We remark that there is a large overlap between the test and train distribution.
This indicates that the discriminator generalizes to the test set and gives evidence against over-fitting.

\cref{fig:toy_z_converge}(b-c) shows the value of $\zeta$ according to different numbers of samples.
Here we see that the computed values converge under a practical number of samples.
Note that since $\zeta$ is a constant of $\dnet$, its computation is required only once and saved for further density estimations.

\begin{figure}
    \centering
    \includegraphics[width=0.9\columnwidth]{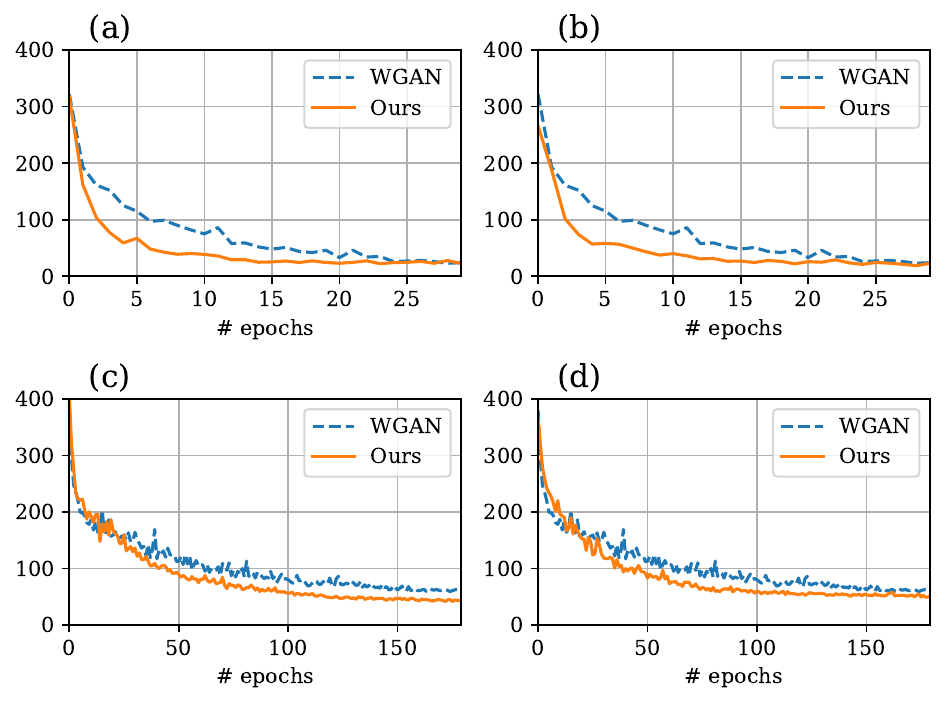}
    \caption{Evolution of FID during training using WGAN loss and our loss. Top row (a-b) for CelebA and bottom row (c-d) for CIFAR-10. The left column (a,c) uses a 1-sample approximation of $\zeta$ and the right column uses a 2-sample approximation.}
    \label{fig:real_fids}
\end{figure}

\subsection{Qualitative results} 
\label{sec:results_qual}

In \cref{fig:celeba_sort,fig:cifar_sort} we visually assess the probabilities assigned by the discriminator by providing random samples sorted by density. We can see here that high-quality images score higher probabilities than low-quality images, suggesting that our trained discriminator has captured the distribution of the given dataset.

\begin{figure}[!ht]
    \centering
    \includegraphics[width=\columnwidth]{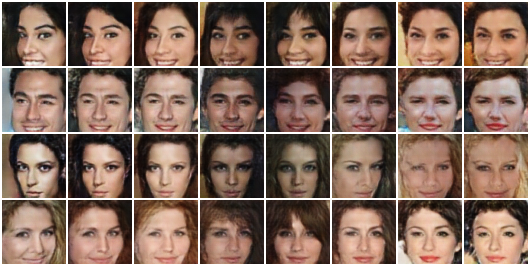}
    \caption{Generated images from linear interpolations of the latent space using the CelebA dataset. Each row is independent of the other.}
    \label{fig:short_interp}
\end{figure}

In \cref{fig:short_interp} %
we show qualitative results for the generator on an interpolation experiment, which shows that our framework retains smoothness in the latent space. 

\begin{figure}[!ht]
    \centering
    \includegraphics[width=\columnwidth]{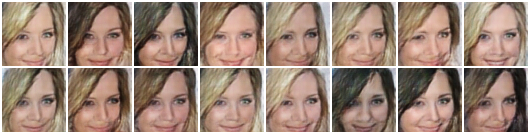}
    \caption{Generator outputs given the same latent vector.}%
    \label{fig:celeba_same_z}
\end{figure}

Finally, because our generator introduces noise when increasing dimensionality (\cref{eq:random_concat}), we wanted to see what characteristics are controlled by the initial latent space.
For that we used the same latent vector as an input to the generator multiple times and show the results in \cref{fig:celeba_same_z}.
We observe that the structure stays the same among the images, with slight variations, \eg hair color.
From here we hypothesize that changing the variance of groups of latent variables can be used as a mechanism to capture qualitative modes of the data.
This corresponds to similar observations in StyleGAN~\cite{karras2019style} where the authors found interesting roles of intermediate auxiliary noise as they were introduced at different resolutions.
However, in contrast here different noise variance plays a crucial role and highlights qualitatively high level of disentanglement in our generator.

\begin{table}[t]
\caption{Comparison of FID values between our model and (1) WGAN with DCGAN architecture, (2) GLOW}
\label{table:FID_score_b_ps}
\begin{center}
\begin{small}
\begin{sc}
\begin{tabular}{lclc}
\hline
                                           & CelebA FID &  & CIFAR-10 FID \\ \hline
\textbf{GLOW}             & 24         &  & 95           \\
\textbf{WGAN-GP}          & 24         &  & 61           \\
\textbf{Ours - 1 sample}  & \textbf{22.5}       &  & \textbf{42.4}         \\
\textbf{Ours - 2 samples} & 22.9       &  & 51.2         \\ \hline
\end{tabular}
\end{sc}
\end{small}
\end{center}
\vspace{0.1in}
\end{table}

\section{Discussion}
\label{LED:sec:Discusion}
As described in~\cref{ssec:gen_loss}, considering a likelihood-based approach for GAN training leads to maximization of the generator entropy in addition to the WGAN objective.
Moreover, the new discriminator objective formulation, as described in~\cref{ssec:model_def}, assists in removing the bias from the WGAN objective.
Both of these differences from the WGAN objectives are made possible by having the generator provide the density of the generated samples.
Whereas normalizing flows require computing the density of arbitrary points in order to train with log-likelihood maximization, a crucial difference with our model is that the computation of the density of real data points is not required. Only the density of the generated data points needs to be computed. %
This leads to a considerable relaxation in the generator architecture, where the model is allowed to increase dimensionality throughout the generator.
The ability to increase dimensions appears to contribute to getting better quality images, as seen in \cref{table:FID_score_b_ps}, where GLOW, which has constant dimensionality, generates lower quality images than DCGAN or more modern GANs.
To keep this operation tractable we adopted %
the approximate Jacobian determinant computation in~\cref{ssec:arch}.
This arguably introduces noise in the gradient.
We leave to future work the task of building a generator architecture with layers that have a closed-form Jacobian. %
For instance, the computation of the Jacobian determinant for a convolution operation can be obtained from~\cite{sedghi2018the,karami2019invertible}, and the Jacobian for element-wise layers is a diagonal matrix. %
We expect this will further speed up and stabilize GAN training.
Furthermore, 
to track the probability of the output of a dimension-reducing layer, as in the final convolution layer of DC-GAN, the removed dimensions have to be marginalized, which is a difficult and expensive computation.
When tractable down-sampling operation is discovered, it could be applied in our model as well.
While experimenting with different number of samples for $\zeta$, we observed that increasing the number of samples did not necessarily improve the image quality.
We suspect that this is because, given more samples, the training focuses more on increasing the variance of the generated images. %
We also suspect that the architecture we used for testing did not have the capacity to accommodate these variances.
We leave it to future work to unlock the full potential of using multiple samples for approximating the normalizing factor.

Finally, we leave to future work the application of the proposed work to more modern GAN generator and discriminator architectures (\eg~\cite{karras2019analyzing}).

\section{Conclusion}
\label{LED:sec:conclusion}
We presented a framework for density estimation within GANs, and explored the connection between EBMs and GANs to develop an unbiased estimator of the partition function of an EBM.
This led to an objective function that is closely related to Wasserstein GAN with an additional entropy maximization criterion for the generator training that enables greater diversity of the generated samples.
Furthermore, we proposed a modified flow network as generator, called one-way flow, which provides both samples and density estimates to compute empirical expectations while maintaining architectural flexibility.
This  allows for an efficient way of evaluating the generator density and generator entropy, which has historically proven hard.
Our experimental results show that our model produces samples that are on par with other GAN generators, along with accurate density estimations and faster convergence.
Our model provides new understandings of the properties of the discriminator and insights into GANs from a maximum likelihood perspective, while connecting these to EBMs. 
To accommodate maximum flexibility we have used a stochastic Jacobian-determinant approximator; we leave as future work its exact computation, which we hypothesize can reduce variance and speed up training.

\subsection*{Acknowledgments}
The authors thank the International Max
Planck Research School for Intelligent Systems for supporting OB. MJB has received research gift funds from Adobe, Intel, Nvidia, Meta/Facebook, and Amazon.  MJB has financial interests in Amazon, Datagen Technologies, and Meshcapade GmbH.  While MJB is a consultant for Meshcapade, his research in this project was performed solely at, and funded solely by, the Max Planck Society.

{\small
\bibliographystyle{ieee_fullname}
\bibliography{references}
}

\newpage
\appendix
\onecolumn
\section{Derivation of objectives}
\subsection{Discriminator objective}
\label{ssec:disc_derivation}
\begin{equation}
    \begin{split}\theta^{*} & =\arg\max_{\theta}\left\{ \sum_{x\in\mathcal{X}}\log P_{D}\left(x\right)\right\} \\
 & =\arg\max_{\theta}\left\{ \sum_{x\in\mathcal{X}}\log\frac{1}{\zeta}e^{D_{\theta}\left(x\right)}\right\} \\
 & =\arg\max_{\theta}\left\{ \sum_{x\in\mathcal{X}}\left(D_{\theta}\left(x\right)-\log\zeta\right)\right\} \\
 & =\arg\max_{\theta}\left\{ \sum_{x\in\mathcal{X}}\left[D_{\theta}\left(x\right)-\log\sum_{y\sim\pg}\frac{e^{\dnet\left(y\right)}}{\pg\left(y\right)}+\log S\right]\right\} \\
 & =\arg\max_{\theta}\left\{ \sum_{x\in\mathcal{X}}\left[D_{\theta}\left(x\right)-\log\sum_{y\sim\pg}e^{\dnet\left(y\right)-\log\pg\left(y\right)}\right]\right\} \\
 & =\arg\max_{\theta}\left\{ \sum_{x\in\mathcal{X}}\left[D_{\theta}\left(x\right)-\log\sum\limits _{z\sim P_{Z}\left(z\right)}e^{D_{\theta}\left(\gnet\left(z\right)\right)-\log P_{Z}\left(z\right)+J_{\gnet}\left(z\right)}\right]\right\} \\
 & =\arg\min_{\theta}\left\{ \sum_{x\in\mathcal{X}}\left[-\log e^{D_{\theta}\left(x\right)}+\log\sum\limits _{z\sim P_{Z}\left(z\right)}e^{D_{\theta}\left(\gnet\left(z\right)\right)-\log P_{Z}\left(z\right)+J_{\gnet}\left(z\right)}\right]\right\} \\
 & =\arg\min_{\theta}\left\{ \sum_{x\in\mathcal{X}}\left[\log\sum\limits _{z\sim P_{Z}\left(z\right)}e^{D_{\theta}\left(\gnet\left(z\right)\right)-D_{\theta}\left(x\right)}e^{J_{\gnet}\left(z\right)-\log P_{Z}\left(z\right)}\right]\right\} 
\end{split}
\end{equation}

Where $J_{\gnet}\left(z\right)=\log\left|\det\left(\frac{\partial\gnet\left(z\right)}{\partial z^{\intercal}}\right)\right|$ is the log Jacobian determinant of $\gnet$ at $z$.

\subsection{Generator objective}
\label{ssec:gen_derivation}
\begin{equation}
\begin{split}\psi^{*} & =\arg\min_{\psi}\text{KL}\left[\pg\left(y\right)\parallel\pd\left(y\right)\right]\\
 & =\arg\min_{\psi}\left\{ \mathbb{E}_{y\sim P_{G}\left(y\right)}\log\left(\frac{\pg\left(y\right)}{\pd\left(y\right)}\right)\right\} \\
 & =\arg\min_{\psi}\left\{ \mathbb{E}_{y\sim P_{G}\left(y\right)}\left[\log\left(\pg\left(y\right)\right)\right]-\mathbb{E}_{y\sim P_{G}\left(y\right)}\left[\log\left(\pd\left(y\right)\right)\right]\right\} \\
\left[\text{Definition of entropy}\right] & =\arg\min_{\psi}\left\{ -H\left(y\right)-\frac{1}{m}\sum_{y\sim P_{G}\left(y\right)}\log\left(\pd\left(y\right)\right)\right\} \\
\left[y=\gnet\left(z\right)\right] & =\arg\min_{\psi}\left\{ -H\left(\gnet\left(z\right)\right)-\frac{1}{m}\sum_{z\sim P_{Z}\left(z\right)}\log\left(\frac{e^{\dnet\left(\gnet\left(z\right)\right)}}{\zeta}\right)\right\} \\
 & =\arg\min_{\psi}\left\{ -H\left(\gnet\left(z\right)\right)-\frac{1}{m}\sum_{z\sim P_{Z}\left(z\right)}\dnet\left(\gnet\left(z\right)\right)\right\} \\
 & =\arg\max_{\psi}\left\{ H\left(\gnet\left(z\right)\right)+\frac{1}{m}\sum_{z\sim P_{Z}}\dnet\left(\gnet\left(z\right)\right)\right\} 
\end{split}
\end{equation}

\section{Optimization using Jacobian approximation} \label{sec:optim_jacobian_effect}
Using a single random sample to approximate the Jacobian determinant as in \cref{eq:det_approx} does not provide an accurate estimate.
Yet, this approximation is effective at maximization of its determinant.
We confirm this claim by optimizing the approximation of the Jacobian and measure the effect on the true Jacobian.

\begin{figure}[ht]
\vspace{0.1in}
\begin{center}
\centerline{\includegraphics[width=\columnwidth]{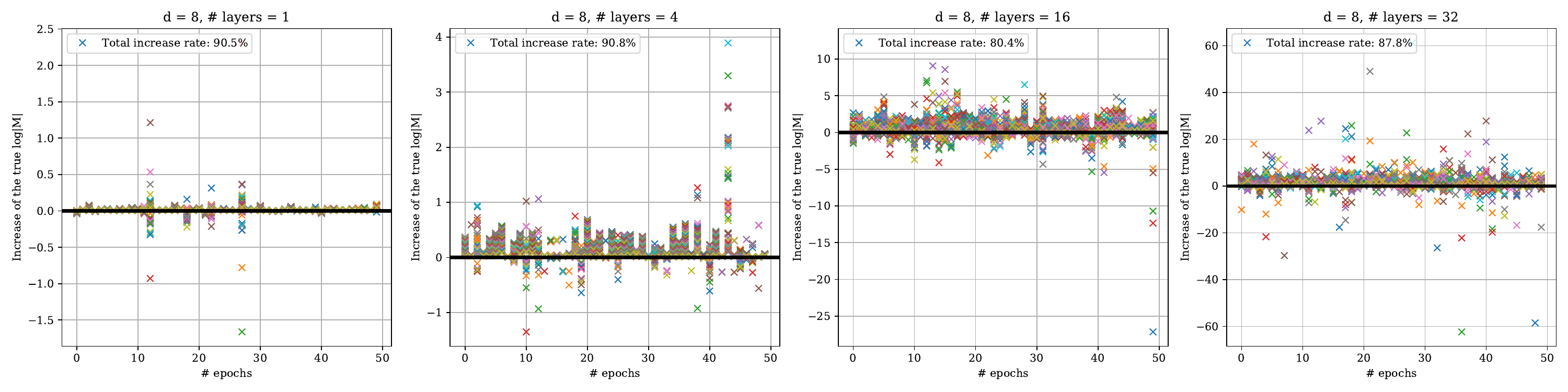}}
\centerline{\includegraphics[width=\columnwidth]{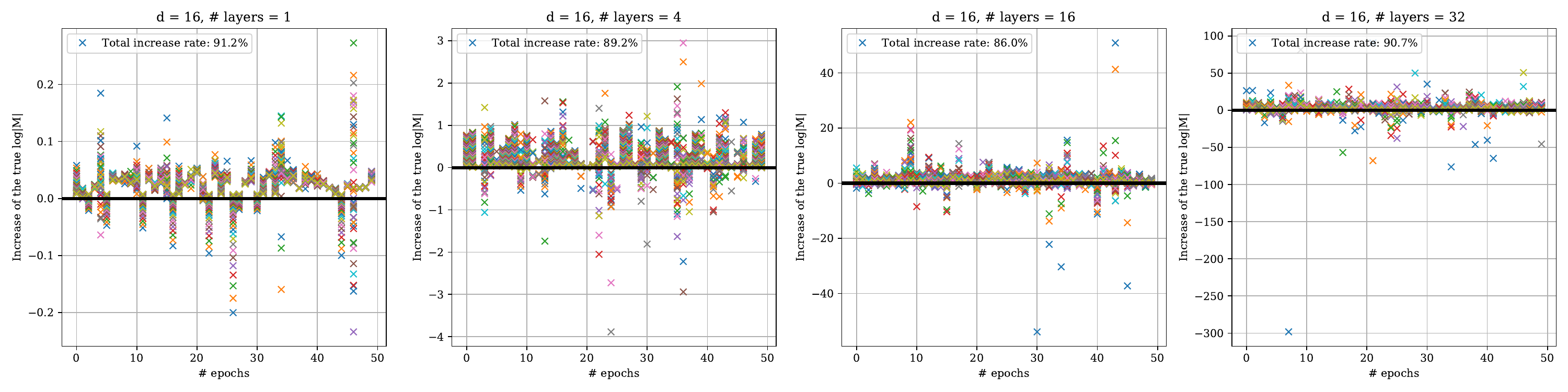}}
\centerline{\includegraphics[width=\columnwidth]{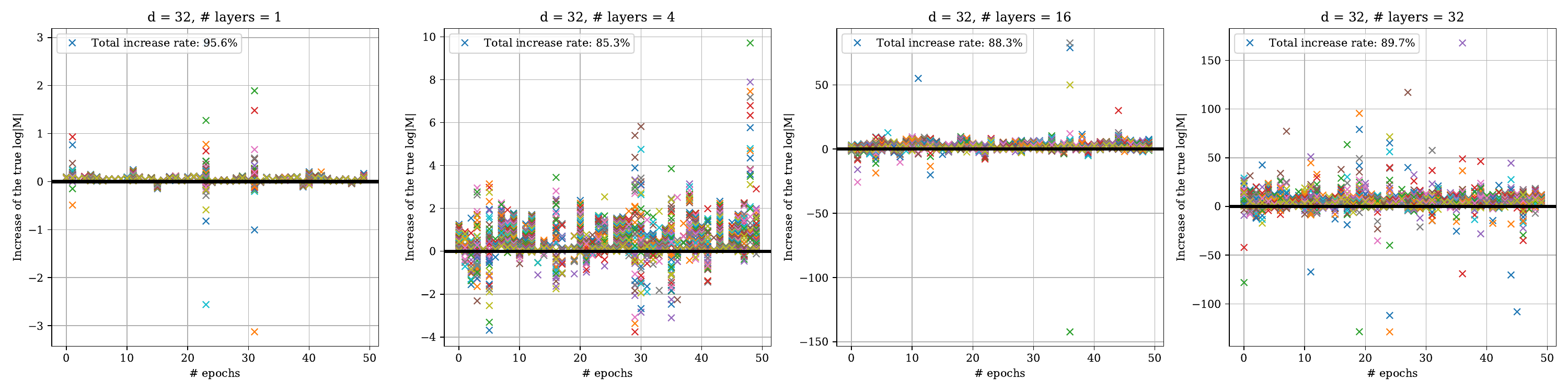}}
\caption{The difference between iterations of the true log determinant of a network trained to maximize the 1-sample approximation of the Jacobian log determinant. Each column represents a different network size and each row corresponds to a different latent size. The black bold line marks 0, where every mark above it means the true value increased.}
\label{fig:ld_approx_maximize}
\end{center}
\vspace{0.1in}
\end{figure}

In this experiment, we randomized 50 neural networks (using a random order of linear, convolution, LeakyRelu or batchnorm layers).
Each network was trained to maximize the approximation of the log determinant of the Jacobian.
After each optimization step, we computed the difference in the value of the log determinant of the true Jacobian from the previous step.
We repeated this experiment with different vector size (8, 16, 32) and different network sizes (1, 4, 16, 32 layers, comparable with DCGAN) with a learning rate of 5e-4.
In \cref{fig:ld_approx_maximize} we mark the differences between consecutive steps of the true log Jacobian.
We define the success rate to be the percentage of times the log determinant increases, and present this metric in each graph.
As can be seen, in all cases the success rate is above 85\%.

\section{Implementation}

\subsection{Numerical stability with large Jacobian determinant}
\label{sec:jacobian_weight}
During training, since the determinant of the Jacobian is generally a different order of magnitude than the discriminator's response (\eg in the objectives in \cref{eq:gen_objective,eq:disc_objective}), it can cause instability in the gradients.
To solve this, we add a scalar $w$ and re-define the probability as
\begin{equation}
    P_{D}\left(y\right)=\frac{e^{\frac{1}{w}D\left(y\right)}}{\zeta}.
\end{equation}
This results in a slightly modified discriminator objective (\cref{eq:disc_objective}):
\begin{equation}
    \begin{split}\theta^{*} & =\arg\max_{\theta}\left\{ \sum_{x\in\mathcal{X}}\log P_{D}\left(x\right)\right\} \\
 & =\arg\max_{\theta}\left\{ \sum_{x\in\mathcal{X}}\log\frac{1}{\zeta}e^{\frac{1}{w}D_{\theta}\left(x\right)}\right\} \\
 & =\arg\max_{\theta}\left\{ \sum_{x\in\mathcal{X}}\left(\frac{1}{w}D_{\theta}\left(x\right)-\log\zeta\right)\right\} \\
 & =\arg\max_{\theta}\left\{ \sum_{x\in\mathcal{X}}\left[\frac{1}{w}D_{\theta}\left(x\right)-\log\sum_{y\sim\pg}\frac{e^{\frac{1}{w}\dnet\left(x\right)}}{\pg\left(y\right)}\right]\right\} \\
 & =\arg\min_{\theta}\left\{ \sum_{x\in\mathcal{X}}\left[-\frac{1}{w}D_{\theta}\left(x\right)+\log\sum\limits _{y\sim P_{G}\left(y\right)}\exp\left(\frac{1}{w}D_{\theta}\left(y\right)-\log\pg\left(y\right)\right)\right]\right\} .
\end{split}
\end{equation}
For the generator, we use a scaled KL-divergence (f-divergence with $f\left(t\right)=wt\log t$.
This results in the modified generator objective (\cref{eq:gen_objective}):
\begin{equation}
\begin{split}\mathcal{D}_{f}\left(P_{G}\parallel P_{D}\right) & =\int f\left(\frac{P_{G}\left(y\right)}{P_{D}\left(y\right)}\right)P_{D}\left(y\right)\text{d}y\\
 & =\int w\frac{P_{G}\left(y\right)}{P_{D}\left(y\right)}\log\left(\frac{P_{G}\left(y\right)}{P_{D}\left(y\right)}\right)P_{D}\left(y\right)\text{d}y\\
 & =\mathbb{E}_{y\sim P_{G}}\left[w\log\left(\frac{P_{G}\left(y\right)}{P_{D}\left(y\right)}\right)\right]\\
 & =\frac{1}{S}\sum_{y\sim P_{G}}\left[w\log\left(P_{G}\left(y\right)\right)-w\log\left(\frac{e^{\frac{1}{w}D\left(y\right)}}{\zeta}\right)\right]\\
 & =\sum_{y\sim P_{G}}\left[-w\log\left|J\left(y\right)\right|-D\left(y\right)\right].
\end{split}
\end{equation}

\subsection{Implementation Details} \label{sec:implemetation_dets}
We trained the generator and discriminator using the PyTorch ADAM optimizer~\cite{adam}.
We set the learning rate of the discriminator to 1e-5 and of the generator to 5e-4.
We trained \modelname{} on the CelebA~\cite{celeba} for 30 epochs and CIFAR-10~\cite{krizhevsky2009learning} for 180 epochs.
We generated images using a latent vector of size 100.

\subsection{Architecture}
\label{sec:architecture}

\subsubsection{Synthetic data network}
For the synthetic data problems in \cref{sec:results_synth_data}, our generator is a 2-layered MLP.
We used the pytorch autograd \emph{jacobian} function to compute the Jacobian and its determinant.

\subsubsection{DC-GAN based generator architecture} \label{sec:dcgan_arch}
The generator architecture we used for CIFAR-10 and CelebA is based on the DC-GAN architecture \cite{radford2015dcgan}.
Given a latent vector, we concatenate a random normal vector.
We then pass this vector to the DC-GAN layers and return the output.
In order to compute $\left\Vert Jv\right\Vert$ from \cref{eq:jacobian_approx}, we use the \emph{jvp} (Jacobian-vector multiplication) function from pytorch.

\subsection{Run times}
\label{ssec:runtime}
To compare the effect of the model on the run time of training, we used a NVIDIA A100-SXM4-40GB GPU for all training on CelebA (64 pixels per edge) and CIFAR-10 (32 pixels per edge).
\cref{table:runtimes} shows the run time per iteration in seconds.
The table shows that applying the one-way flow almost doubles the time per iteration, but using a few additional samples does not increase the run time significantly further.

\begin{table}
\caption{Comparison of time per training iteration in seconds between our model and WGAN with DCGAN architecture. The time is formatted as $\left\langle \text{mean}\right\rangle \pm\left\langle \text{standard deviation}\right\rangle $}
\label{table:runtimes}
\begin{center}
\begin{sc}
\begin{tabular}{lclc}
\hline
                           & CelebA &                & CIFAR-10 \\ \hline
\textbf{WGAN-GP}           &  $0.031 \pm  0.00072$  &  & $0.025 \pm  0.00067$  \\
\textbf{Ours - 1 sample}   &  $0.053 \pm  0.00181$  &  & $0.044 \pm  0.00078$  \\
\textbf{Ours - 2 samples}  &  $0.063 \pm  0.00183$  &  & $0.045 \pm  0.00231$  \\
\hline
\end{tabular}
\end{sc}
\end{center}
\vspace{0.1in}
\end{table}

\end{document}